%% file: main.tex
\documentclass{llncs}

\usepackage{eccv}
\usepackage{stfloats}
\usepackage{float}
\usepackage{eccvabbrv}
\usepackage{graphicx}
\usepackage{booktabs}
\usepackage[accsupp]{axessibility}
\usepackage{xcolor}
\definecolor{myblue}{RGB}{0, 51, 102}
\usepackage[colorlinks,
            linkcolor = myblue,
            urlcolor  = myblue,
            citecolor = myblue,
]{hyperref}
\usepackage{orcidlink}
\newcommand{\jy}[1]{}

\begin{document}

\title{LightHarmony3D: Harmonizing Illumination and Shadows for Object Insertion in 3D Gaussian Splatting}

\titlerunning{LightHarmony3D}

\author{Tianyu Huang\inst{1}$^{*}$ \and
Zhenyang Ren\inst{1}$^{*}$ \and
Zhenchen Wan\inst{1} \and
Jiyang Zheng\inst{1} \and
Wenjie Wang\inst{2} \and
Runnan Chen\inst{1} \and
Mingming Gong\inst{2} \and
Tongliang Liu\inst{1}}

\authorrunning{T.~Huang et al.}

\institute{The University of Sydney, Sydney, Australia \and
The University of Melbourne, Melbourne, Australia\\
$^{*}$Equal contribution.}

\makeatletter
\apptocmd{\@maketitle}{\vspace{-1em}
   \centering
   \includegraphics[width=0.9\textwidth]{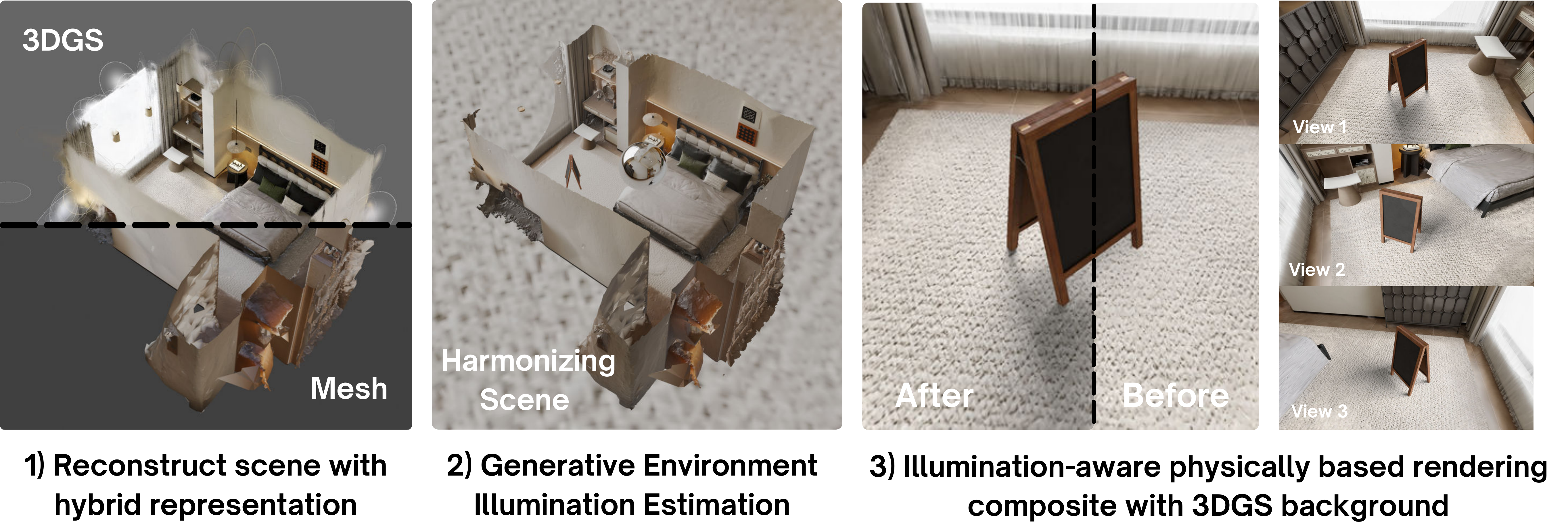}
   \vspace{-0.5em}
   \captionof{figure}{We propose \textbf{LightHarmony3D}, a novel framework for illumination-consistent insertion of external meshes into 3D Gaussian Splatting scenes. A hybrid Gaussian-mesh representation captures scene structure, diffusion-based panorama prediction recovers dominant lighting as an HDR environment map, and PBR integrates the inserted mesh with the original 3DGS renderings to produce coherent multi-view results.}
   \label{fig:teaser}
}{}{}
\makeatother

\maketitle

\input{sec/0_abstract}

\input{sec/1_intro}
\input{sec/2_related_work}
\input{sec/3_method}
\input{sec/4_experiment}
\input{sec/5_conclusion}

\clearpage
\bibliographystyle{splncs04}
\bibliography{main}
\end{document}

%% file: sec/0_abstract.tex
\begin{abstract}
3D Gaussian Splatting (3DGS) enables high-fidelity reconstruction of scene geometry and appearance. Building on this capability, inserting external mesh objects into reconstructed 3DGS scenes enables interactive editing and content augmentation for immersive applications such as AR/VR, virtual staging, and digital content creation. However, achieving physically consistent lighting and shadows for mesh insertion remains challenging, as it requires accurate scene illumination estimation and multi-view consistent rendering. To address this challenge, we present LightHarmony3D, a novel framework for illumination-consistent mesh insertion in 3DGS scenes. Central to our approach is our proposed generative module that predicts a full 360° HDR environment map at the insertion location via a single forward pass. By leveraging generative priors instead of iterative optimization, our method efficiently captures dominant scene illumination and enables physically grounded shading and shadows for inserted meshes while maintaining multi-view coherence. Furthermore, we introduce the first dedicated benchmark for mesh insertion in 3DGS, providing a standardized evaluation framework for assessing lighting consistency and photorealism. Extensive experiments across multiple real-world reconstruction datasets demonstrate that LightHarmony3D achieves state-of-the-art realism and multi-view consistency.
\end{abstract}

%% file: sec/1_intro.tex
\section{Introduction}
Recent progress in 3D Gaussian Splatting (3DGS)~\cite{kerbl_3d_2023} has revolutionized novel view synthesis, enabling high-fidelity, real-time scene reconstruction. This makes it a compelling representation for immersive 3D applications such as augmented reality, virtual reality, and digital content creation. In many of these applications, users frequently need to modify reconstructed scenes by inserting new virtual objects for interactive editing and content augmentation. In practical graphics pipelines, virtual assets are typically represented as explicit polygon meshes due to their precise geometry, high-frequency surface detail, and well-established material representations~\cite{akenine2019real}. To support real-world editing workflows, it is therefore essential that such mesh-based assets can be seamlessly inserted into reconstructed 3DGS scenes with physically consistent lighting and shadows.

Achieving photorealistic mesh insertion in 3DGS scenes remains significantly challenging due to the critical need to maintain consistent illumination and physical shadow interactions with the surrounding environment.  Unlike traditional graphics pipelines where lighting is explicitly represented, 3DGS implicitly entangles illumination and material properties within view-dependent Gaussian colors, making scene lighting difficult to recover~\cite{kerbl_3d_2023}. Existing approaches attempt to address this either through inverse rendering~\cite{gao_relightable_2024, chen_gi-gs_2025}, which requires expensive optimization and accurate geometry, or through generative HDR prediction methods such as GaSLight~\cite{gaslight}, which estimate lighting independently for each view but often suffer from inconsistent illumination across viewpoints. Beyond lighting estimation, integrating an explicit mesh with a volumetric Gaussian scene further complicates visibility reasoning and shadow compositing~\cite{chen_gi-gs_2025,guedon_milo_2025, lin2025iris}. Without accurate geometry-aware interaction between the inserted mesh and the Gaussian field, naive compositing often leads to mismatched shading or implausible shadow artifacts, breaking visual realism.

To address these challenges, we present LightHarmony3D, a framework for illumination-consistent insertion of explicit mesh objects into 3DGS scenes. Instead of relying on computationally expensive inverse rendering, we estimate scene illumination through a generative module that predicts low-exposure, light-source-centric panoramas at the target insertion location using a fine-tuned diffusion model~\cite{ho2020denoising}. The generated panoramas are fused into a full 360$^\circ$ HDR environment map, enabling efficient recovery of dominant scene illumination in a single forward pass. To achieve physically consistent shadow integration, we further introduce a PBR-guided shadow compositing strategy. Leveraging a hybrid Gaussian–mesh representation~\cite{guedon_milo_2025}, the scene is rendered in a physically based rendering (PBR) engine~\cite{shirley2009fundamentals} to compute a shadow ratio map that captures mesh-scene interactions. This map is then used to modulate the original 3DGS rendering, allowing physically grounded cast shadows to be injected while preserving the photorealistic appearance of the reconstructed scene.

By synergizing geometric reconstruction, generative illumination priors, and shadow-aware compositing, LightHarmony3D achieves photorealistic mesh insertion inside 3DGS environments, addressing a key missing capability for digital content pipelines. Our main contributions are summarized as follows:
\begin{itemize}
\item We introduce LightHarmony3D, a framework for physically consistent mesh insertion into 3DGS scenes, successfully bridging the gap between standard explicit 3D assets and 3DGS-style representations.
\item We propose GenEnvLighting, a generative illumination module that synthesizes accurate HDR environment maps via diffusion priors, circumventing the need for complex and resource-intensive inverse rendering.
\item We design a PBR-Guided Shadow Compositing technique that utilizes shadow ratio modulation to seamlessly integrate cast shadows onto the 3DGS background without compromising the original scene's visual fidelity.
\item We establish a new benchmark with high-quality synthetic scenes for mesh-based object insertion in 3DGS, demonstrating that LightHarmony3D achieves robust lighting consistency, multi-view coherence, and photorealism.
\end{itemize}

%% file: sec/2_related_work.tex
\section{Related Works}

\subsection{3DGS and Surface Extraction}
While Neural Radiance Fields (NeRFs)~\cite{mildenhall2021nerf} pioneered continuous volumetric scene representations, 3D Gaussian Splatting (3DGS)~\cite{kerbl_3d_2023} has emerged as a highly efficient explicit alternative, enabling real-time rendering of high-fidelity scenes. However, native 3DGS is inherently point-based and lacks the explicit topological surfaces required for accurate physical simulations. To bridge this gap, early methods explored implicit neural representations such as neural SDFs~\cite{wang2021neus} or hybrid Gaussian-implicit frameworks~\cite{huang20242d, chen2023neusg, chen2024vcr}, though they often inherit the optimization bottlenecks of MLPs. 
Other works attempt to extract meshes post-optimization~\cite{guedon2024sugar, yu2024gaussian}, yet naive isosurfacing frequently introduces structural artifacts in thin or high-frequency regions. In contrast, recent differentiable frameworks like MILo~\cite{guedon_milo_2025} integrate meshing directly into the 3DGS optimization loop using Delaunay triangulation. Our pipeline leverages these advancements, utilizing the co-optimized explicit mesh as a reliable geometric foundation for downstream physically based light transport.

\subsection{Scene Editing and Inverse Rendering}
A rapidly expanding body of literature explores editing within 3DGS. Early methods leverage 2D segmentation models~\cite{kirillov2023segment} to isolate Gaussians belonging to individual objects~\cite{cen_segment_2024, ye_gaussian_2024, choi_click-gaussian_nodate}. More recent approaches incorporate vision-language models for semantic-aware manipulation~\cite{wu_opengaussian_2024, qin_langsplat_2024, shi2024language, zhou_feature_2024, huang2025openinsgaussian}. Complementary attribute editing frameworks~\cite{chen_gaussianeditor_2023, lee2025editsplat, wu2024gaussctrl} allow for localized modifications of Gaussian colors or opacities. While highly expressive, these techniques operate exclusively within the Gaussian parameter space and lack the mechanisms to physically integrate external mesh assets. To achieve realistic lighting, inverse rendering approaches~\cite{chen_gi-gs_2025, gao_relightable_2024, liang2024gs} decompose 3DGS into geometry, materials, and incident illumination. Despite producing physically grounded relighting, these methods demand computationally prohibitive per-scene optimization and dense material estimation. More recently, concurrent works attempt to bypass heavy inverse rendering via 2D generative priors. MV-CoLight~\cite{ren_mv-colight_2025} lifts harmonized 2D composites to 3DGS, but its feed-forward image-domain operations preclude accurate physical mesh integration. Similarly, GaSLight~\cite{gaslight} employs diffusion models to predict HDR lighting on a per-view basis; however, this discrete view-level prediction inherently struggles to maintain the strict spatial consistency required for coherent global illumination.

\subsection{Generative Illumination Estimation}
Diffusion models~\cite{ho2020denoising} implicitly encode rich, natural lighting priors learned from vast image corpora. Recent works have successfully exploited these priors for illumination estimation. For instance, DiPIR~\cite{liang_photorealistic_2024} combines personalized diffusion with inverse rendering to recover scene lighting, while DiffusionLight~\cite{phongthawee_diffusionlight_2024} reconstructs HDR environment maps by inpainting a chrome ball, revealing high-intensity light sources difficult to infer from LDR imagery alone. Our work, LightHarmony3D, draws inspiration from these generative capabilities but fundamentally reorients them for 3DGS object insertion. Instead of relying on cumbersome inverse rendering pipelines or per-view approximations, we fine-tune a diffusion model to directly predict a 360$^\circ$ low-exposure panorama at the target location. By coupling this generative HDR illumination with the extracted explicit mesh within a standard physically based rendering pipeline, we achieve realistic, geometry-aware, and multi-view consistent mesh insertion without necessitating costly optimization or large paired compositing datasets. 

%% file: sec/3_method.tex
\section{Method}
\subsection{3D Gaussian and Surface Reconstruction}
\begin{figure*}[t]
\centering
\includegraphics[width=\textwidth]{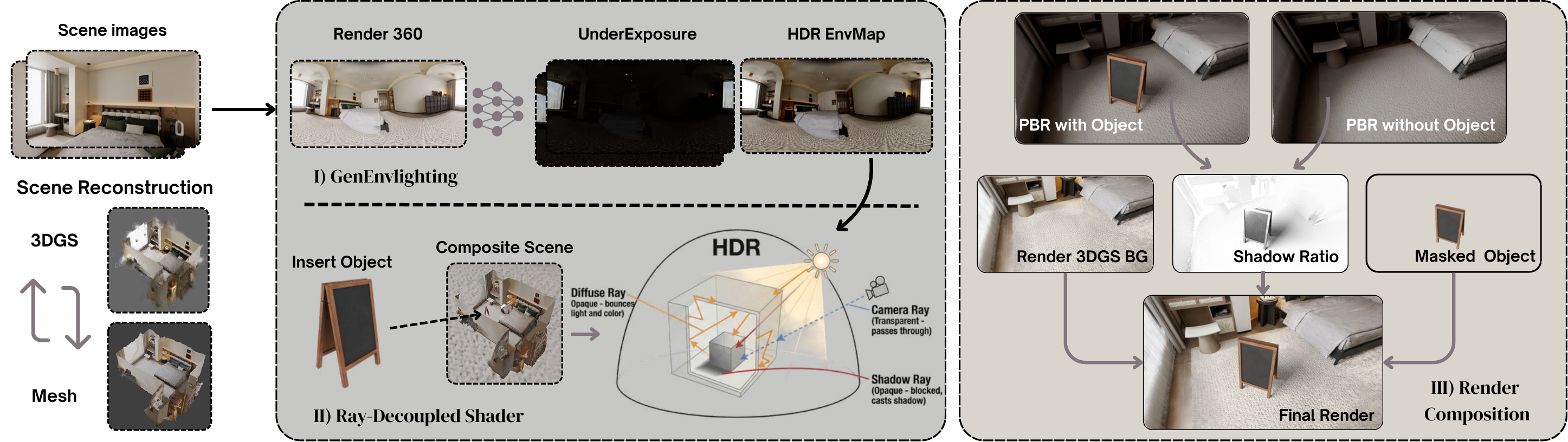}
\caption{\textbf{Overview of LightHarmony3D.} Given multi-view input images, we reconstruct a hybrid 3DGS–mesh scene and render a base 360$^\circ$ panorama. To capture dominant light emitters, a fine-tuned diffusion model predicts bracketed underexposures, which are fused to reconstruct a HDR environment map via radiometric truncation. To illuminate the enclosed scene, we employ a ray-decoupled visibility formulation that renders the proxy mesh transparent to camera rays while preserving its physical shadow-catching properties. Finally, we perform physically based rendering of the inserted mesh and linearly composite the resulting physically grounded shading and cast shadows over the original 3DGS background to produce seamless, illumination-consistent results.}
\label{fig:model}
\end{figure*}
Our pipeline begins with reconstructing a full 3D scene from input images using MILo~\cite{guedon_milo_2025}. MILo leverages the 3D Gaussian Splatting representation and differentiably extracts a triangle mesh during training, enforcing consistency between the volumetric Gaussians and the surface mesh. This approach produces a high-fidelity textured mesh of the scene with an order of magnitude fewer vertices than previous methods, making it efficient for downstream rendering. The resulting mesh preserves fine geometric details and captures the complete scene structure (foreground objects and background) in a single unified model. We will use this mesh both for environment lighting estimation and for physically correct shadow rendering in later steps.

\subsection{Environment Map Generation from Reconstructed Scene}
To capture local illumination, we generate a base 360$^\circ$ environment map by rendering six orthogonal Low Dynamic Range (LDR) perspective views centered at the target insertion point within the reconstructed 3DGS scene. These views are subsequently stitched into a unified equirectangular panorama. Denoted as $\text{EV}_0$, this base panorama faithfully preserves the ambient radiance and spatial color distribution of the original environment.

Beyond providing the baseline ambient lighting for the inserted object, the $\text{EV}_0$ panorama serves as the conditioning input for our subsequent generative illumination module. Rather than relying on heuristic semantic extraction, we formulate the identification of dominant light emitters as a physical radiometric truncation process. As detailed in the next section, our diffusion model utilizes this $\text{EV}_0$ input to simulate a bracketed underexposure regime (e.g., $\text{EV}_{-3}$). Under this simulated underexposure, surfaces with lower radiant exitance naturally attenuate below the clipping threshold, which elegantly isolates high-energy illumination directions based purely on radiometric properties.

\subsection{Generative Environment Illumination Estimation}
Following the extraction of the base-exposure ($\text{EV}_0$) environment map, we reframe the identification of dominant lighting as an exposure-controlled radiometric truncation task. We fine-tune a latent diffusion model to map standard-exposure panoramas to a simulated reduced exposure regime (e.g., $\text{EV}_{-3}$). Our objective is to generate an underexposed representation where surfaces with lower radiant exitance-such as diffuse background reflections-naturally attenuate below the visibility threshold. This radiometric clipping effectively isolates high-energy illumination directions without requiring explicit semantic segmentation of light sources. We build upon Flux.1 Kontext~\cite{batifol2025flux}, a pretrained diffusion model whose native image-conditioned editing capability makes it well-suited for this global exposure transformation.

Given the relatively limited size of our dataset (800+ HDR panoramas collected from Poly Haven~\cite{polyhaven}), directly fine-tuning the full model risks overfitting. We therefore adopt a DreamBooth-style training strategy~\cite{ruiz2023dreambooth} combined with LoRA (Low-Rank Adaptation)~\cite{hu2022lora}. DreamBooth enables consistent learning of the task-specific transformation concept—mapping normal exposure to a radiometrically truncated exposure—while LoRA constrains adaptation to low-rank updates, significantly reducing the number of trainable parameters.

For training data synthesis, each HDR environment map is tone-mapped to multiple exposure levels. The $\text{EV}_0$ panorama serves as the input condition, and its mathematically underexposed counterpart ($\text{EV}_{-3}$) is used as the supervision target. The $\text{EV}_0$ image is encoded into the latent space via the model’s VAE encoder, while a fixed text prompt guides the model to perform the exposure reduction. During fine-tuning, the base model weights are frozen and only the injected LoRA parameters are updated:
$$\Delta W_i = A_i B_i, \quad W'_i = W_i + \alpha \, \Delta W_i,$$
where $A_i$ and $B_i$ are learnable parameters, $\alpha$ is a scaling factor, and $\text{rank}(A_i) = d \ll \text{dim}(W_i)$. 

After fine-tuning, the model robustly generalizes to unseen environment maps, producing physically plausible underexposed outputs where dominant radiant energy is preserved while secondary bounces are attenuated.

\subsection{HDR Environment Map Construction via Exposure Fusion}
To recover a high-dynamic-range (HDR) representation of the scene illumination, we fuse the base-exposure panorama with a bracketed sequence of underexposed panoramas (e.g., down to $\text{EV}_{-6}$). Since the generated darker panoramas may exhibit minor structural generative variations compared to the deterministic input, standard pixel-wise HDR merging algorithms can introduce severe ghosting artifacts. Our goal is to robustly restore clipped high-intensity radiant energy across a wide dynamic range while strictly maintaining the stable chromatic content of the original 3DGS scene. To this end, we adopt a bottom-up, luminance-based iterative merging strategy~\cite{phongthawee_diffusionlight_2024}, which prioritizes radiometric consistency over exact pixel-wise alignment.

Given a sequence of $N$ LDR panoramas $\{I_i\}_{i=1}^N$ corresponding to exposure values $\{\text{EV}_i\}_{i=1}^N$ arranged in ascending order (e.g., $\text{EV}_1 = -6, \text{EV}_2 = -3, \text{EV}_3 = 0$), we explicitly distinguish their origins: the base exposure $I_N$ is directly rendered from the reconstructed 3DGS, whereas the underexposed sequence $\{I_i\}_{i=1}^{N-1}$ is sequentially predicted by our fine-tuned diffusion model. 

We first compute the normalized luminance of each image in linear space:
\begin{equation}
\tilde{L}_i(p) = \boldsymbol{w}^\top \big(I_i(p)\big)^\gamma,
\end{equation}
where $\gamma = 2.4$ performs inverse gamma correction to linearize the sRGB inputs, and $\boldsymbol{w} = [0.21267,\; 0.71516,\; 0.07217]$ is the standard RGB-to-luminance weighting vector. The corresponding absolute luminance (or relative irradiance) for each exposure level is then scaled to a unified radiometric space:
\begin{equation}
E_i(p) = \tilde{L}_i(p) \cdot 2^{-\text{EV}_i}.
\end{equation}

To recover the expanded dynamic range, we iteratively merge these estimates from the darkest exposure ($i=1$) upwards to the base exposure ($i=N$). Let the initial fused radiance be $E'_1(p) = E_1(p)$. For $i = 2, \dots, N$, we update the fused luminance sequentially:
\begin{equation}
E'_i(p) = 
\begin{cases} 
E'_{i-1}(p), & \text{if } \tilde{L}_i(p) > 0.9 \\ 
E_i(p), & \text{otherwise.}
\end{cases}
\end{equation}
This formulation cleanly replaces saturated regions in the brighter image—where the normalized luminance $\tilde{L}_i(p)$ exceeds the $0.9$ threshold—with the radiometrically consistent, unclipped values from the darker images. In practice, these discrete replacements are blended using a smooth spatial transition mask near the threshold boundary to suppress visual discontinuities.

Finally, the HDR environment map is reconstructed by applying the fully merged luminance $E'_N$ back to the chromaticity of the original base image $I_N$:
\begin{equation}
\text{HDR}(p) = \frac{E'_N(p)}{\tilde{L}_N(p)} \cdot \big(I_N(p)\big)^\gamma.
\end{equation}

By utilizing a sequence of bracketed exposures, this targeted luminance replacement efficiently recovers a broad dynamic range (e.g., 6 stops). The resulting 32-bit floating-point EXR panorama robustly encodes both ambient illumination and extremely high-intensity light sources, yielding the wide radiometric range essential for accurate physically based shadow casting.

\subsection{Ray-Decoupled Visibility for Enclosed Scene Illumination}
A fundamental challenge in applying image-based lighting (IBL) to reconstructed scenes is the topological closure of the extracted geometry. Explicit meshes derived from 3DGS or SDFs often form continuous, watertight enclosures without explicitly modeled architectural openings (like windows or doors). Consequently, directly assigning the HDR environment map as the world illumination results in complete light occlusion, rendering the interior of the scene unphysically dark.

Rather than resorting to heuristic mesh culling or modifying the underlying topology, we introduce a Ray-Decoupled Visibility Formulation. By leveraging the path-tracing mechanics of PBR engines, we design a custom light-path conditional shader for the reconstructed scene mesh. This formulation selectively decouples illumination penetration from shadow reception based on the specific type of the evaluated ray.

Mathematically, let $\omega$ denote an evaluated ray in the path tracer, and let $\tau(\omega)$ represent its semantic ray type. We define a binary indicator function $\mathcal{I}(\omega)$ to classify the interaction:
\begin{equation}
\mathcal{I}(\omega) = 
\begin{cases} 
1, & \text{if } \tau(\omega) \in \{\text{Shadow}, \text{Diffuse}\} \\ 
0, & \text{if } \tau(\omega) \in \{\text{Camera}, \text{Transmission}, \text{Glossy}\}
\end{cases}
\end{equation}

For any surface point $p$ on the reconstructed scene mesh, the effective Bidirectional Scattering Distribution Function (BSDF), denoted as $f(p, \omega_i, \omega_o)$, is computed as a linear interpolation between a physically-based opaque material and a perfectly transparent medium, governed by the incoming ray type $\tau(\omega_i)$:
\begin{equation}
f(p, \omega_i, \omega_o) = \mathcal{I}(\omega_i) f_{\text{Principled}}(p, \omega_i, \omega_o) + \big(1 - \mathcal{I}(\omega_i)\big) f_{\text{Transparent}}(p, \omega_i, \omega_o)
\end{equation}
where $f_{\text{Principled}}$ is parameterized by the reconstructed vertex colors to model accurate diffuse transport, and $f_{\text{Transparent}}$ allows light to pass through without interaction or refraction.

This ray-decoupled mechanism elegantly solves the occlusion paradox. Because camera and environment rays ($\mathcal{I} = 0$) perceive the mesh as transparent, the external HDR illumination can effortlessly penetrate the enclosed scene boundaries to illuminate the interior and the inserted virtual object. Simultaneously, when the path tracer evaluates shadow rays or indirect diffuse bounces ($\mathcal{I} = 1$), the mesh correctly acts as a solid geometric receiver. 

\subsection{Physically-Guided Compositing and Shadow Refinement}

Given the recovered HDR environment map, we render both the reconstructed receiver mesh and the virtual object using a path tracer (Blender Cycles~\cite{blender}). By assigning the HDR panorama as the world illumination, we ensure that the object and the mesh receive consistent diffuse and specular lighting. For each viewpoint, we generate two renders of the receiver: a reference image $R_0$ (without the object) and an image $R_1$ (with the object present). Additionally, we render the isolated object image $O$ (with an alpha channel) and extract the corresponding 3DGS background $B$. The 3DGS image retains the high-frequency appearance of the original scene without synthetic shadows, serving as the pristine background to avoid artifacts of double shadows.

\noindent\textbf{Linear-Space Shadow Estimation.} 
To physically quantify the shading attenuation induced by the inserted object, we compute a per-pixel shadow ratio map. To accurately model colored lighting and physically grounded light transport, it is imperative to estimate this ratio independently for each RGB channel in the linear color space, rather than relying on a monochromatic luminance approximation. Let $\phi(\cdot)$ and $\phi^{-1}(\cdot)$ denote the standard conversions between the sRGB and linear color spaces. We first map the renders of the receiver to the linear space: $\tilde{R}_0 = \phi(R_0)$ and $\tilde{R}_1 = \phi(R_1)$. For each color channel $c \in \{r,g,b\}$, we compute the raw shadow ratio as:
\begin{equation}
S_c(p) = 
\begin{cases}
\mathrm{clip}\!\left(\dfrac{\tilde{R}_{1,c}(p)}{\tilde{R}_{0,c}(p)+\epsilon},\,0,\,1\right), & p \in \mathcal{V}_c,\\
1, & \text{otherwise},
\end{cases}
\end{equation}
Here, $\epsilon$ represents a small constant for numerical stability. The term $\mathcal{V}_c$ defines a validity mask that gates unreliable ratios in extremely dark regions by enforcing a minimum energy bound on $\tilde{R}_{0,c}$, thereby preventing the amplification of noise.

\noindent\textbf{Parametric Shadow Shaping.} 
Directly applying the raw ratio $S$ often produces shadows that are overly sharp or excessively deep, which are caused by minor geometric mismatches between the proxy mesh and the 3DGS representation. To mitigate this issue, we refine the ratio via a parametric shaping function applied to each channel:
\begin{equation}
\tilde{S}_c(p) = \max\!\big(S_c(p)^{\gamma},\, s_{\min}\big), \qquad \hat{S}_c(p) = 1 - \lambda \big(1 - \tilde{S}_c(p)\big),
\end{equation}
where $\gamma \in (0,1]$ modulates the contrast (a smaller $\gamma$ yields softer falloffs of shadows), $s_{\min}$ acts as a lower-bound clamp to prevent unnaturally pure black regions, and $\lambda \in [0,1]$ scales the global intensity of the shadow. These parameters offer robust control over the profile of the shadow and are typically kept fixed across scenes.

\noindent\textbf{Compositing with 3DGS Background.} 
Finally, we apply the refined, per-channel shadow ratio $\hat{S}$ to the 3DGS background in the linear space and composite the rendered object. Let $M(p) \in [0,1]$ denote the object mask derived from the alpha channel of $O$. We formulate the linear composite $\tilde{C}$ as:
\begin{equation}
\tilde{C}(p) = \big(\tilde{B}(p) \odot \hat{S}(p)\big)\,\big(1 - M(p)\big) \;+\; \tilde{O}(p)\,M(p),
\end{equation}
where $\tilde{B}=\phi(B)$, $\tilde{O}=\phi(O)$, and $\odot$ denotes element-wise multiplication. The first term modulates the 3DGS background with the physically simulated, colored cast shadows, while the second term overlays the object illuminated by the HDR map. We then obtain the final sRGB output via $C = \phi^{-1}(\tilde{C})$. This refinement ensures that the injected shadows are radiometrically consistent with the illumination of the scene, effectively handling complex colored lighting while preserving the photorealism of the original 3DGS reconstruction.

%% file: sec/4_experiment.tex
\section{Experiment}
\begin{figure}[t]
    \centering
    \includegraphics[width=\linewidth]{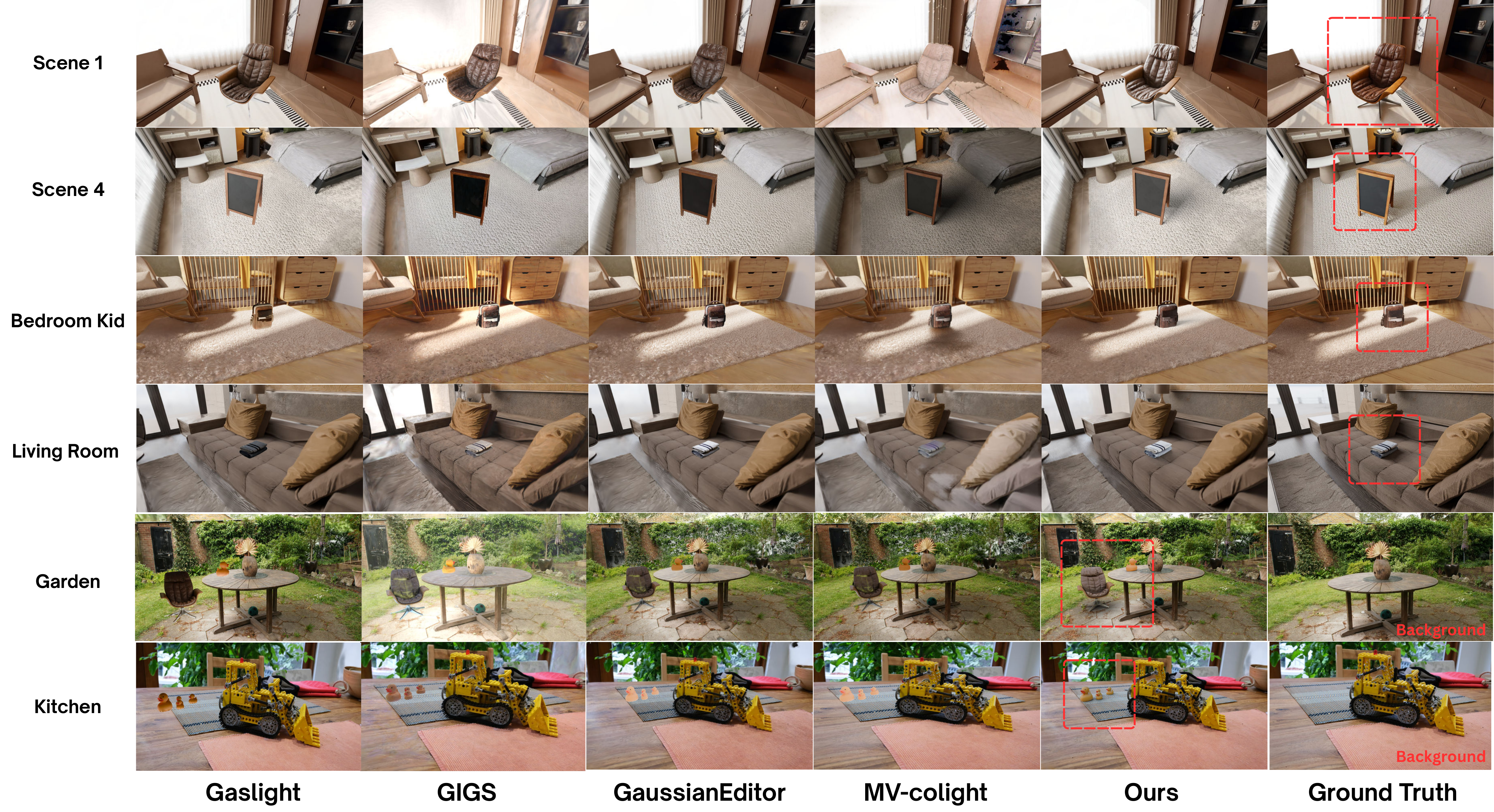}
    \caption{Qualitative results on our LH3D synthetic dataset and real scenes from Mip-NeRF360. The inserted object is highlighted with a bounding box. As Mip-NeRF360 provides no insertion ground truth, we additionally show the original background for reference.}
    \label{fig:Qualitative}
\end{figure}

\subsection{Experimental Setup}
\noindent\textbf{Datasets.}
We evaluate on both real-world and synthetic data. For real-world scenes we use Mip-NeRF360~\cite{barron2022mip}, which offers diverse indoor and outdoor environments with high-quality multi-view imagery. Quantitative evaluation of mesh insertion in 3DGS demands pixel-exact ground-truth pairs where the inserted object receives physically correct shading and casts view-consistent shadows under the scene's global illumination—paired supervision absent from all existing benchmarks, as each image must be path-traced with and without the object under identical lighting to capture even subtle shadow colour shifts and indirect-light contributions.
\input{tab/tab3}

We therefore construct LH3D-Bench through a fully manual pipeline. For each scene we design the environment and lighting, place objects at physically plausible locations, craft camera trajectories optimised for 3DGS training, and path-trace paired ground-truth images at high sample counts. We also develop a Blender-to-COLMAP export plugin that directly produces the initial point cloud and camera parameters for 3DGS initialisation from the Blender scene, which will be released alongside the data. On average, a single scene requires $\sim$12\,h of manual work (placement, trajectory design, data export) and $\sim$16 GPU-hours for paired rendering; the design stages cannot be automated, as placement plausibility and lighting representativeness require human judgement.

The benchmark spans two subsets: LH3D-Ku (4 Kujiale~\cite{Kujiale} indoor scenes with PolyHaven~\cite{polyhaven} objects, 20 test pairs) and LH3D-Blender (7 BlenderKit scenes $\times$ 9 objects, 315 test images), incorporating multi-source lighting and diverse material properties to stress-test generalization.

\noindent\textbf{Evaluation Metrics.} 
For the synthetic benchmark with ground truth, we measure PSNR, SSIM~\cite{wang2004image}, and LPIPS~\cite{zhang2018lpips} on the rendered multi-view results to assess illumination consistency and perceptual fidelity. Because each ground-truth image represents a unique, physically plausible combination of lighting and material properties, PSNR alone is insufficient to capture perceptual realism. Therefore, SSIM and LPIPS are critical for evaluating structural integrity and perceptual quality in the regions of the inserted objects. Furthermore, we pioneer a reference-free evaluation protocol for object insertion in the absence of ground truth by leveraging a Vision-Language Model. Specifically, we employ VQAScore~\cite{lin2024evaluating} to formulate a contrastive metric that evaluates the realism and harmonization of the insertion. As a vision-language alignment metric, VQAScore provides a robust perceptual signal for assessing visual plausibility without requiring reference images. For each synthesized image, we compute alignment scores against a positive prompt (describing a natural, well-lit insertion with consistent shadows) and a negative prompt (describing inconsistent lighting, halos, and unnatural shadows). We compute the ratio between the positive and negative scores, defined as 
$\text{Ratio} = \frac{\text{pos}}{\text{pos} + \text{neg}}$, 
which provides a robust and automated measure of photorealism without requiring reference images.

\noindent\textbf{Baselines.} 
We compare our approach with four representative methods that cover the main paradigms related to our task. Specifically, we include a Gaussian-based inverse-rendering approach (GIGS~\cite{chen_gi-gs_2025}), a 3DGS editing framework capable of object-level scene modification (GaussianEditor~\cite{chen_gaussianeditor_2023}), a multi-view compositing method that lifts harmonized 2D edits to 3DGS (MV-CoLight~\cite{ren_mv-colight_2025}), and a concurrent generative illumination approach that predicts HDR lighting on a per-view basis (GaSLight~\cite{gaslight}). These baselines collectively cover inverse rendering, Gaussian-space editing, multi-view compositing, and generative illumination estimation, but only GaSLight natively supports inserting external meshes. For a fair comparison, we adapt each baseline by reconstructing every object into 3DGS from 128 dense spherical-orbit views so that performance gaps reflect the lighting pipeline rather than input degradation. All data and tools will be publicly released.

\subsection{Performance Analysis}
\noindent\textbf{Quantitative Results.}
As shown in Tab.~\ref{tab:lh3d_results} and Tab.~\ref{tab:blender_results}, LightHarmony3D achieves the highest scores across most quantitative metrics on both the LH3D-Ku and LH3D-Blender benchmarks, demonstrating strong illumination consistency and high-fidelity compositing. On the LH3D-Ku dataset, the concurrent method GaSLight performs reasonably but falls short of our framework due to its inconsistent per-view predictions. GaussianEditor ranks third overall; its pipeline depends on 2D inpainting and coarse image-to-3D reconstruction, followed by direct Gaussian concatenation, which lacks both physically grounded relighting and robust visibility reasoning. GIGS performs notably worse overall because its environment map is fixed to the original scene geometry, causing strong over-exposure after insertion. MV-CoLight shows the weakest absolute fidelity, as its feed-forward harmonization predicts color adjustments rather than modeling physical illumination. On the LH3D-Blender dataset (Tab.~\ref{tab:blender_results}), our method maintains its overall superiority with the highest PSNR and SSIM, proving its robustness across diverse geometric configurations.

\begin{figure}[t]
\centering
\includegraphics[width=0.95\columnwidth]{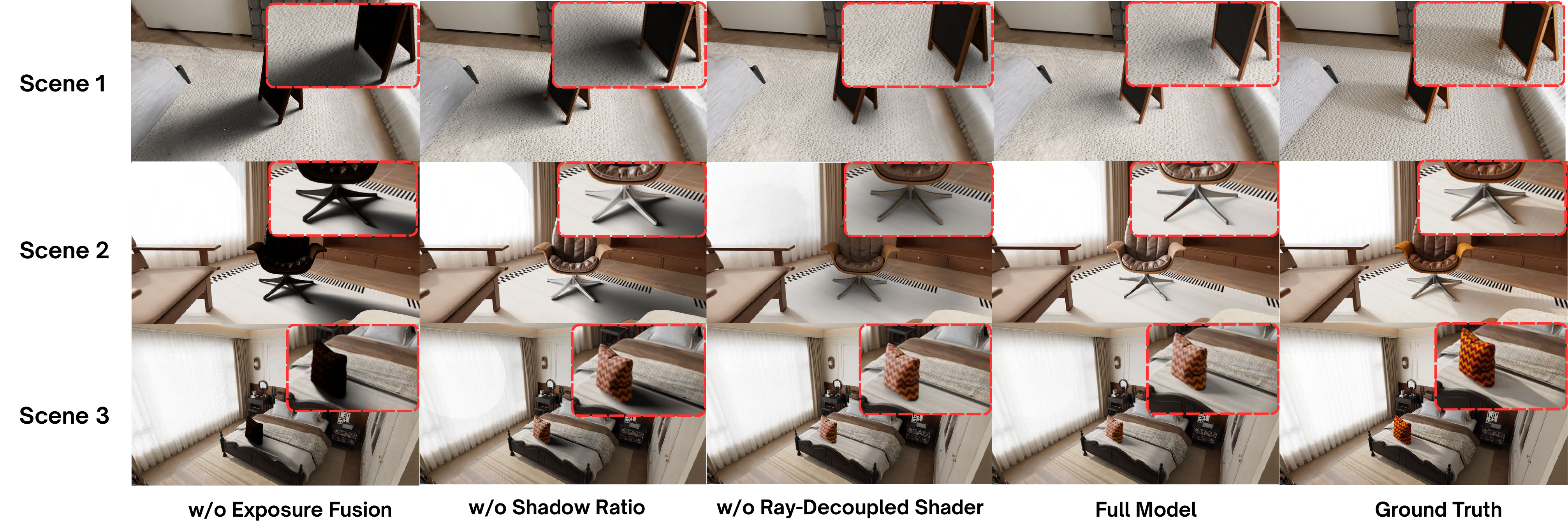}
\caption{Visual results of the ablation study on LH3D-Ku. Details are zoomed in within the bounding boxes.}
\label{fig:Ablation}
\end{figure}
\input{tab/tab1}

Furthermore, to quantitatively assess realism on the real-world Mip-NeRF360 dataset where no ground-truth insertions are available, we employ the VQA-based contrastive evaluation (Tab.~\ref{tab:mipnerf360_vqa}). LightHarmony3D achieves the highest positive alignment score (0.528) and the lowest negative score (0.208), culminating in a state-of-the-art naturalness ratio of 0.751. This confirms that our physically grounded pipeline generates the most perceptually harmonious results without visible compositing artifacts, substantially outperforming all baselines.

\input{tab/tab2}

\noindent\textbf{Qualitative Results.}
Fig.~\ref{fig:Qualitative} presents visual comparisons on the synthetic datasets and real-world Mip-NeRF360 scenes. GIGS often generates overly bright or glossy objects because its lighting representation cannot adapt to new geometry and lacks proper shadow attenuation. GaussianEditor produces objects that visually "float" in the scene, as its geometry receives no physically accurate shading and does not cast shadows on surrounding surfaces. MV-CoLight struggles to generalize due to a significant domain gap in scene complexity. Trained primarily on object-centric tabletop environments with simplistic lighting, its feed-forward harmonization fails to infer the holistic lighting context of our room-scale scenes featuring intricate global illumination and diverse occluders. Consequently, its predictions break down, resulting in overly soft, desaturated shading and missing cast shadows.

For the concurrent baseline GaSLight, we observe two critical failure modes (see Fig.~\ref{fig:Qualitative}). First, its reliance on per-view HDR prediction leads to severe multi-view inconsistencies when evaluated on dense-view reconstructions. This spatial instability actively degrades the underlying quality of the optimized Gaussians, introducing noticeable rendering artifacts. Second, lacking a holistic understanding of global illumination, GaSLight frequently misinterprets brightly illuminated local surfaces as primary light emitters. For instance, when a bag is placed on a sunlit carpet, GaSLight erroneously predicts the bright carpet surface as the light source in certain views, resulting in inverted lighting directions (e.g., illuminating the object from below) and a complete absence of physically correct cast shadows.

In contrast, LightHarmony3D achieves coherent lighting and realistic cast shadows by combining generative radiometric truncation, 360$^\circ$ HDR reconstruction, and ray-decoupled PBR rendering. Inserted objects align naturally with the global scene illumination, preserve correct contact shadows, and maintain a consistent appearance across viewpoints, demonstrating strong generalization to challenging real-world environments.

\subsection{Ablation Study and Extensibility}
As demonstrated in Fig.~\ref{fig:Ablation} and Tab.~\ref{tab:ablation}, each component of LightHarmony3D is essential for achieving physically consistent object insertion. Removing the multi-exposure fused HDR environment map (i.e., relying solely on the base LDR panorama) results in uniformly dim lighting, where the inserted objects appear flat and cast shadows lose their physically correct intensity and directionality. Eliminating the linear-space shadow-ratio formulation and relying on direct alpha-compositing of the raw PBR shadow mask causes the resulting shadows to appear unphysically dark and introduces jarring double-shadowing artifacts over the original 3DGS background. Furthermore, without the ray-decoupled visibility module, the reconstructed watertight proxy mesh acts as a complete occluder, entirely blocking the external HDR illumination. Conversely, naively culling the ceiling geometry to allow light penetration compromises the structural integrity of the scene, thereby failing to generate valid cast shadows for the inserted objects. In contrast, the full LightHarmony3D pipeline yields coherent lighting, accurate cast shadows, and seamless harmonization with the reconstructed scene, demonstrating the complementary indispensability of all proposed modules.

In Fig.~\ref{fig:Extensibility}, we present multi-view insertion results. Our PBR-based rendering framework ensures highly consistent lighting and shadow behavior across viewpoints, and as shown in Fig.~\ref{fig:Qualitative} garden and kitchen, inserting multiple objects does not introduce mutual interference-each object receives scene-consistent illumination, and the shadows remain coherent and physically plausible. We further show consecutive frames from an animation sequence with inserted objects. Because physically based light transport is precomputed, subsequent insertions and local object animations can be generated efficiently while maintaining stable illumination and temporally consistent shadows.

%% file: tab/tab3.tex
\begin{table}[t]
\centering

\begin{minipage}{0.48\linewidth}
\centering
\caption{Quantitative comparison on the LH3D-Ku synthetic dataset.}
\begin{tabular}{lccc}
\toprule
Method & PSNR$\uparrow$ & SSIM$\uparrow$ & LPIPS$\downarrow$ \\
\midrule
GIGS & 17.33 & 0.698 & 0.334 \\
GaussianEditor & 21.56 & 0.812 & 0.215 \\
MV-CoLight & 15.99 & 0.747 & 0.256 \\
GasLight & 20.41 & 0.812 & 0.224 \\
\midrule
Ours & \textbf{24.03} & \textbf{0.832} & \textbf{0.200} \\
\bottomrule
\end{tabular}
\label{tab:lh3d_results}
\end{minipage}
\hfill
\begin{minipage}{0.48\linewidth}
\centering
\caption{VQA score comparison on the Mip-NeRF360 dataset.}
\begin{tabular}{lccc}
\toprule
Method & Positive$\uparrow$ & Negative$\downarrow$ & Ratio$\uparrow$ \\
\midrule
3DGS & 0.351 & 0.400 & 0.501 \\
GasLight & 0.472 & 0.541 & 0.457 \\
GIGS & 0.261 & 0.276 & 0.576 \\
MV-CoLight & 0.339 & 0.421 & 0.387 \\
\midrule
Ours & \textbf{0.528} & \textbf{0.208} & \textbf{0.751} \\
\bottomrule
\end{tabular}
\label{tab:mipnerf360_vqa}
\end{minipage}

\end{table}

%% file: tab/tab1.tex
\begin{table}[t]
\centering
\scriptsize
\setlength{\tabcolsep}{3pt}
\begin{minipage}{0.45\linewidth}
\centering
\caption{Quantitative comparison on the LH3D-Blender dataset.}
\begin{tabular}{lccc}
\toprule
Method & PSNR$\uparrow$ & SSIM$\uparrow$ & LPIPS$\downarrow$ \\
\midrule
GIGS & 20.110 & 0.690 & 0.406 \\
GaussianEditor & 20.680 & 0.696 & 0.357 \\
MV-CoLight & 19.257 & 0.642 & \textbf{0.321} \\
GasLight & 19.171 & 0.626 & 0.399 \\
\midrule
LH3D (Ours) & \textbf{23.987} & \textbf{0.744} & 0.335 \\
\bottomrule
\end{tabular}
\label{tab:blender_results}
\end{minipage}
\hfill
\begin{minipage}{0.48\linewidth}
\centering
\caption{Ablation study of different components in our compositing pipeline.}

\begin{tabular}{lccc}
\toprule
Method & PSNR$\uparrow$ & SSIM$\uparrow$ & LPIPS$\downarrow$ \\
\midrule
w/o Exposure Fusion & 23.198 & 0.8173 & 0.2190 \\
w/o Shadow Ratio & 22.490 & 0.8217 & 0.2093 \\
w/o Ray-Decoupled \\ Shader & 23.543 & 0.8317 & 0.2007 \\
\midrule
Full Model & \textbf{24.032} & \textbf{0.8318} & \textbf{0.2004} \\
\bottomrule
\end{tabular}

\label{tab:ablation}
\end{minipage}

\end{table}

%% file: tab/tab2.tex
\begin{figure}[t]
    \centering
    \includegraphics[width=0.95\columnwidth]{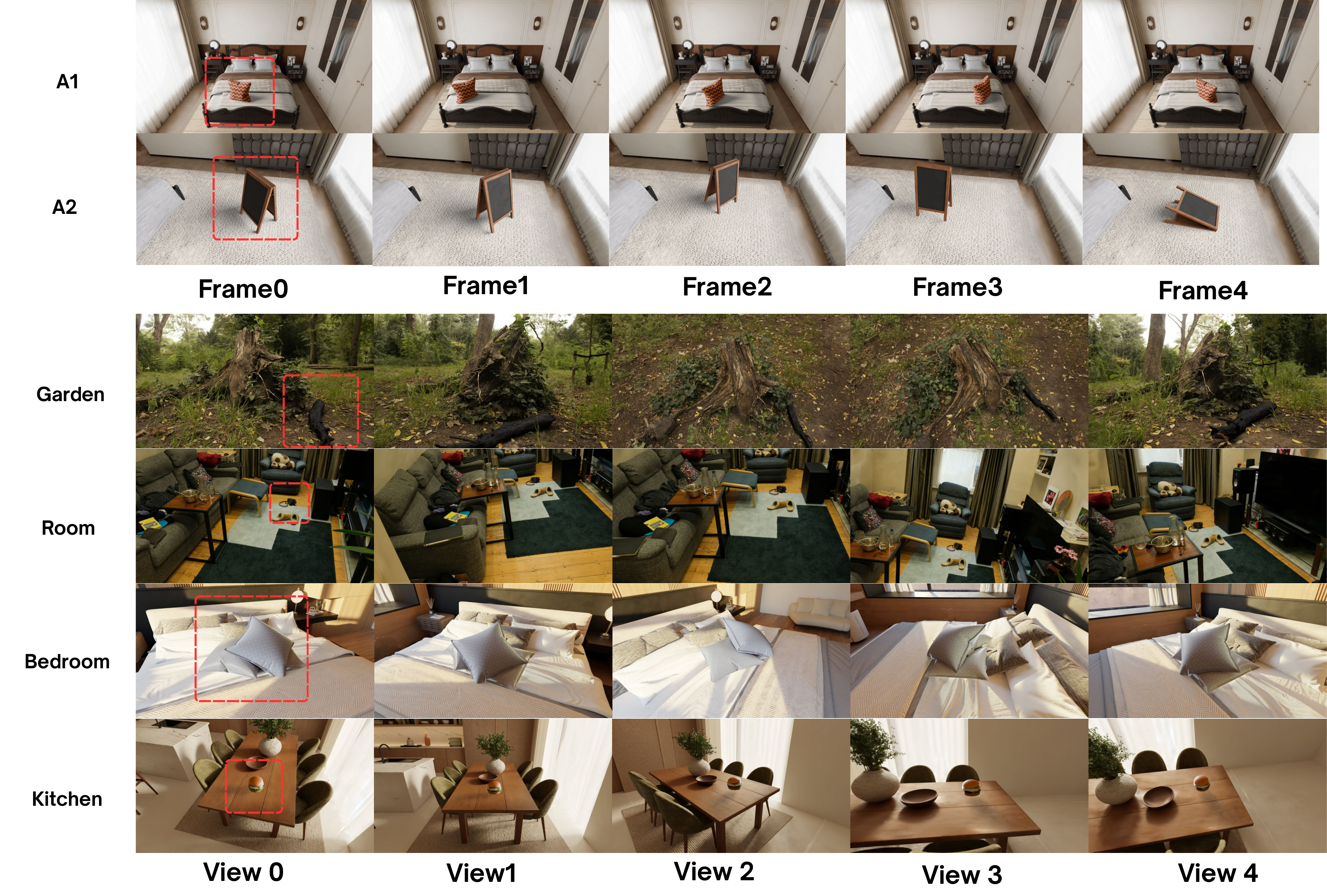}
    \caption{Extensibility. Top: frames from two animation sequences (A1, A2) showing stable illumination across time. Bottom: multi-view insertion results demonstrating consistent shading and shadows from different viewpoints.}
    \label{fig:Extensibility}
\end{figure}

%% file: sec/5_conclusion.tex
\section{Conclusion}
We presented LightHarmony3D, a unified framework for physically consistent mesh insertion into 3DGS-reconstructed scenes. By combining hybrid Gaussian–mesh reconstruction, diffusion-based illumination estimation, and HDR-guided physically based rendering, our method achieves state-of-the-art lighting consistency, realistic shading, and coherent multi-view compositing across diverse scenes. We further introduced a new benchmark to enable systematic evaluation of mesh-insertion quality in 3DGS environments. Despite its strong performance, LightHarmony3D has several limitations. The quality of illumination estimation and shadow synthesis depends on the accuracy of the reconstructed mesh; geometric artifacts or missing structures may lead to suboptimal lighting or incorrect occlusion reasoning. Our diffusion-based light extraction also assumes sufficient coverage of light sources in the reconstructed environment; scenes with extremely sparse observations or complex indirect lighting remain challenging. In addition, the physically based rendering stage introduces non-negligible computational cost compared to purely feed-forward harmonization methods. Future work could explore end-to-end differentiable illumination models, joint optimization of Gaussians and mesh lighting, and learning-based PBR priors to further improve robustness and efficiency.